\documentclass[sigconf, screen]{acmart}
\acmSubmissionID{9371}

\usepackage{booktabs} % For formal tables
\usepackage{multirow}
\usepackage[table]{xcolor}

\acmConference[MM '26]{Proceedings of the 34th ACM International Conference on Multimedia}{November 10--14, 2026}{Rio de Janeiro, Brazil}

\copyrightyear{2026}
\acmYear{2026}
\setcopyright{cc}
\setcctype{by}
\acmConference[MM '26]{Proceedings of the 34th ACM International Conference on Multimedia}{November 10--14, 2026}{Rio de Janeiro, Brazil}
\acmBooktitle{Proceedings of the 34th ACM International Conference on Multimedia (MM '26), November 10--14, 2026, Rio de Janeiro, Brazil}
\acmDOI{10.1145/3767308.3836536}
\acmISBN{979-8-4007-2213-4/2026/11}

\acmDOI{10.1145/3767308.3836536}

\begin{document}
% Title portion
\title{ACA-GS: Adaptive-Capacity Anchored Gaussian Splatting for Compact Dynamic Radiance Fields}

\begin{teaserfigure}
    \centering
    \includegraphics[width=1.0\textwidth]{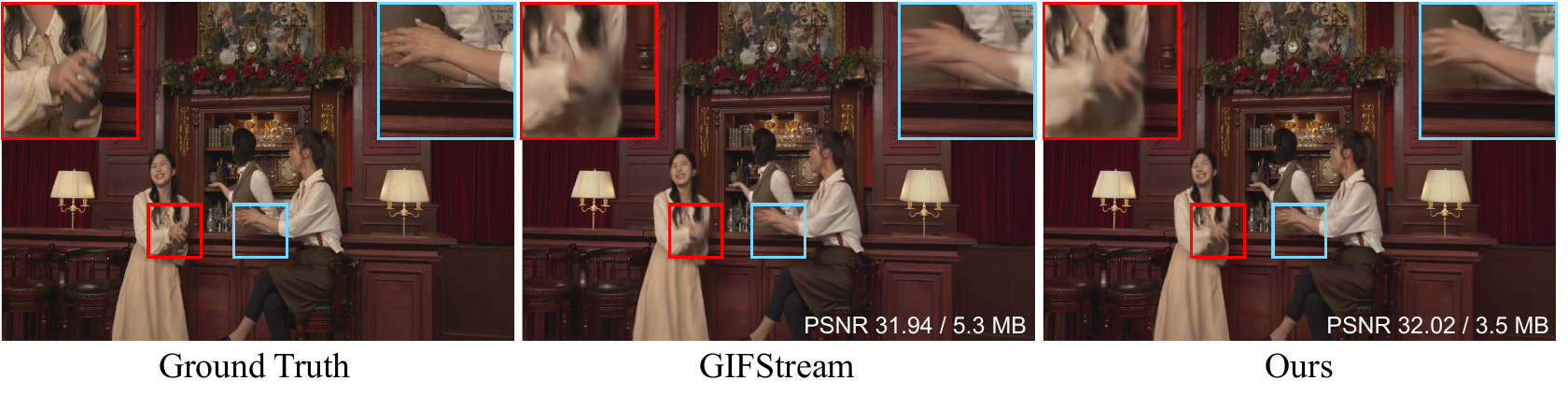}
    \vspace{-2.5em}
    \caption{We propose adaptive-capacity anchor-based representation that dynamically adjusts both per-anchor cardinality and feature capacity, achieving high rendering quality with small storage size at dynamic fast motion scenes.}
\label{fig:demo}
\end{teaserfigure}

\author{Seunghyeon Song}
\orcid{0009-0005-1073-5558}
\affiliation{
 \institution{Sungkyunkwan University}
 \city{Suwon}
 \state{Gyeonggi-do}\\
 \country{Republic of Korea}}
\email{shs4489@skku.edu}

\author{Joo Chan Lee}
\orcid{0000-0001-9398-9089}
\affiliation{
 \institution{Sungkyunkwan University}
 \city{Suwon}
 \state{Gyeonggi-do}\\
 \country{Republic of Korea}}
\email{maincold2@skku.edu}

\author{Chanung Park}
\orcid{0009-0004-9150-1317}
\affiliation{
 \institution{Sungkyunkwan University}
 \city{Suwon}
 \state{Gyeonggi-do}\\
 \country{Republic of Korea}}
\email{pcw980420@g.skku.edu}

\author{Jun Young Jeong}
\orcid{0000-0002-2457-1647}
\affiliation{
 \institution{Electronics and Telecommunications Research Institute}
 \city{Daejeon}
 \country{Republic of Korea}}
\email{jyj0120@etri.re.kr}

\author{Minseo Lee}
\orcid{0009-0007-2379-4084}
\affiliation{
 \institution{Yonsei University}
 \city{Seoul}
 \country{Republic of Korea}}
\email{lms@yonsei.ac.kr}

\author{Eunbyung Park}
\orcid{0000-0003-4071-2814}
\affiliation{
 \institution{Yonsei University}
 \city{Seoul}
 \country{Republic of Korea}}
\email{epark@yonsei.ac.kr}

\author{Jong Hwan Ko}
\authornote{Corresponding author.}
\orcid{0000-0003-4434-4318}
\affiliation{
 \institution{Sungkyunkwan University}
 \city{Suwon}
 \state{Gyeonggi-do}\\
 \country{Republic of Korea}}
\email{jhko@skku.edu}

\renewcommand\shortauthors{Song et al.}

\begin{abstract}
Recent advances in 4D Gaussian Splatting (4DGS) enable high-fidelity, real-time spatiotemporal rendering, but expose a fundamental trade-off between motion expressiveness and storage efficiency. While anchor-based designs achieve compactness through anchor-level parameter sharing, their rigid uniform parametrization enforces fixed Neural Gaussian counts and feature budgets per anchor. Consequently, insufficient fidelity is addressed by excessive anchor density, rather than lightweight, targeted increases in Neural Gaussian count or feature capacity, resulting in memory waste. To overcome this rigidity, we introduce an adaptive-capacity anchor-based framework that dynamically allocates the representational capacity based on local spatiotemporal demands. Adaptive Anchor Cardinality varies the number of Neural Gaussians per anchor, concentrating primitives in regions of high geometric or motion complexity while suppressing redundancy. In parallel, Adaptive Anchor Feature Masking modulates anchor-level feature channels, assigning rich features to complex regions and lightweight representations to simpler ones. Experiments on MPEG, Panoptic Sports, and N3DV datasets demonstrate substantial storage reduction without degrading visual quality. Notably, on challenging MPEG sequences with complex motion, our method achieves up to 1.5× higher compression than state-of-the-art anchor-based methods while preserving comparable quality.
\end{abstract}





\maketitle

\section{Introduction}

Building on the success of neural radiance fields~\cite{mildenhall2020nerf}, 3D Gaussian Splatting (3DGS)~\cite{kerbl3Dgaussians} has emerged as a paradigm-shifting method in novel view synthesis, achieving high-fidelity 3D scene reconstruction with real-time rendering by utilizing explicit 3D Gaussian primitives. Following the success of 3DGS, there has been a surge of research extending this technique from static scenes to the spatiotemporal domain for the effective representation of dynamic scenes. These efforts include training separate 3DGS models per frame~\cite{luiten2023dynamic}, combining a single static model with deformation networks~\cite{bae2024ed3dgs, yang2023deformable3dgs}, or integrating spatiotemporal grid structures~\cite{Wu_2024_CVPR} and 4D rotation/scaling parameters~\cite{Li_STG_2024_CVPR, yang2023gs4d}.

However, these methods face an inherent trade-off between the complexity of dynamic scenes and storage efficiency. Deformation-based methods~\cite{bae2024ed3dgs, yang2023deformable3dgs,Wu_2024_CVPR} show limitations in precisely representing regions with rapid motion, which inevitably leads to a degradation in rendering quality. Conversely, 4D primitive-based methods~\cite{Li_STG_2024_CVPR, yang2023gs4d} attempt to mitigate these artifacts and maintain high quality by narrowing the frame interval or increasing temporal dimensions. However, this results in an exponential increase in storage requirements. 

\begin{figure}[t]
    \centering
    \includegraphics[width=0.9\linewidth]{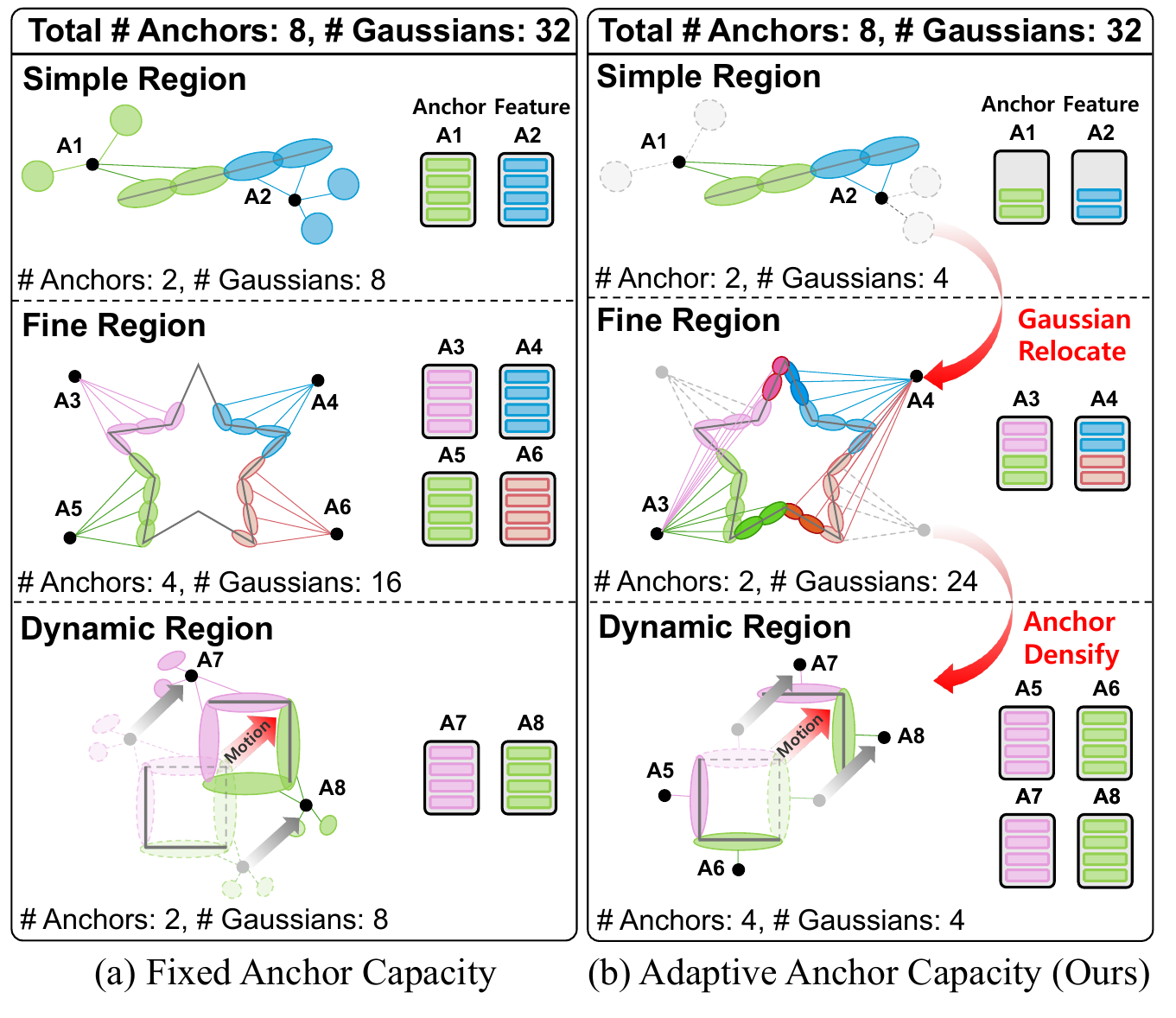}
\vspace{-1em}    
\caption{Overview. (a) Fixed Anchor Capacity enforces uniform Gaussian counts and feature dimensions regardless of local complexity, causing redundancy in simple areas and insufficient detail in complex and dynamic regions. (b) Adaptive Anchor Capacity dynamically scales resources to spatiotemporal demands, eliminating redundancy while ensuring high fidelity in fine and large motion regions.}    
\label{fig:overview}
\vspace{-1em}  
\end{figure}

More recently, several methods extended anchor-based 3D representation, originally proposed in Scaffold-GS~\cite{scaffoldgs}, to dynamic scenes. 
For instance, ADC-GS~\cite{ijcai2025p132} maintained a canonical anchor structure and refined anchors based on temporal significance, and GIFStream~\cite{li2025gifstream4dgaussianbasedimmersive} augmented the anchor representation with a feature stream for temporal prediction, improving robustness under large motions. Despite their promising storage efficiency, these methods inherit a fundamental structural rigidity: each anchor is parameterized with 1) a fixed number of Neural Gaussians (i.e., cardinality) and 2) a uniform anchor feature channel. Given that cardinality and feature dimensions are the primary determinants of an anchor's representational capacity, an identical assignment of these parameters irrespective of local scene complexity results in suboptimal distribution of model resources.

For instance, fine-detail regions or intricate geometry require a high density of Neural Gaussians and sufficient feature capacity to achieve high-fidelity reconstruction. However, when these resources are fixed per anchor, the model must introduce an excessive number of anchors to reach the required Gaussian count, as illustrated in Fig.~\ref{fig:overview}(a). In contrast, in simple backgrounds or low-frequency regions, fixed allocation produces redundant primitives and unnecessarily large feature capacity, resulting in significant storage inefficiency.

In dynamic regions, where predicting individual temporal trajectories for each Gaussian is challenging, anchor-based methods represent motion via anchor-level deformation. By extension, this approach requires dense anchor placement in areas exhibiting large or complex motion. When anchors are sparsely placed in such regions, the model fails to capture rich motion patterns, leading to noticeable quality degradation, as illustrated in Fig.~\ref{fig:overview}(a). Conversely, increasing anchor density to capture such motion can waste capacity by introducing redundant primitives due to the fixed Neural Gaussian cardinality. Furthermore, anchor feature capacity should be adaptively modulated based on local spatiotemporal complexity to ensure robust representational power for both intricate geometry and diverse motion patterns.

\begin{figure}[t]
    \centering
    \includegraphics[width=0.9\linewidth]{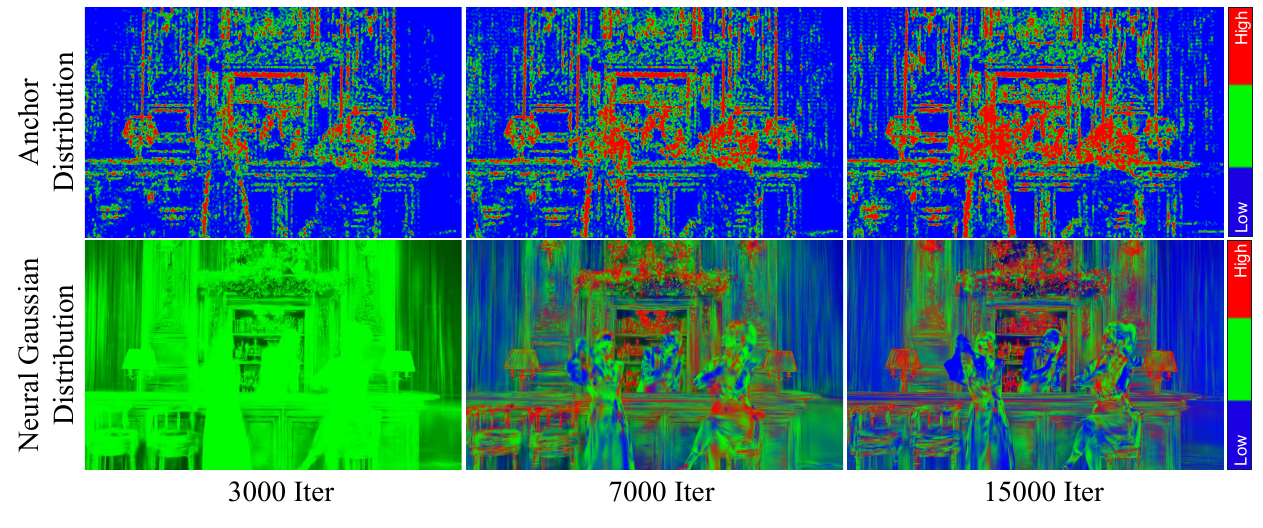}
    \vspace{-1em}
    \caption{Distribution of Anchor and Neural Gaussian within our method over training iterations. Per-pixel anchor counts: Blue (0--1), Green (2--4), Red ($\ge$5). Per-anchor Neural Gaussians: Blue (1--3), Green (4--6), Red (7--10).}
\label{fig:iterAD}
\vspace{-1em}  
\end{figure}

In this work, we propose an adaptive anchor representation for dynamic scenes, which can adjust 1) the Neural Gaussian cardinality and 2) the anchor feature channels for each anchor (Fig.~\ref{fig:overview}(b)). 
First, for adaptive cardinality, we propose an importance-based relocation strategy that identifies unnecessary Neural Gaussians and relocates them to important regions, binding them to nearby anchors.
To measure the importance of primitives from each anchor, we introduce an accumulated view-dependent opacity metric, since prior anchor-based methods do not explicitly represent per-primitive view-independent opacity.
Based on this importance measure, Neural Gaussians are adaptively reallocated according to spatial and dynamic complexity, assigning higher Gaussian density to fine regions while keeping simple regions compact without redundant primitives (Fig.~\ref{fig:iterAD}).

Second, instead of utilizing the entire feature channel of an anchor, we introduce learnable channel masks that selectively activate spatial features according to local complexity.
Aggressively reducing spatial features for motion-critical anchors can induce spatio-temporal interference, where temporal capacity is consumed by spatial details. We therefore introduce a conditional regularization loss to preserve sufficient feature capacity in high-motion regions.

We evaluated the proposed framework on various dynamic novel view synthesis datasets, including N3DV~\cite{li2022neural3dvideosynthesis}, Panoptic Sports~\cite{joo2016panopticstudiomassivelymultiview}, and MPEG datasets. The experimental results demonstrate that our method achieves high rendering quality with compact storage size. Notably, on the MPEG dataset, which includes fast and large motions, our model achieved approximately $\mathbf{1.5\times}$ higher compression rates compared to the previous state-of-the-art method~\cite{li2025gifstream4dgaussianbasedimmersive} while maintaining perceptual quality. This validates that our adaptive resource allocation approach is an effective solution for achieving high-quality rendering and storage efficiency in dynamic 3D scenes.

\begin{figure*}[t]
    \centering
    \includegraphics[width=0.9\textwidth]{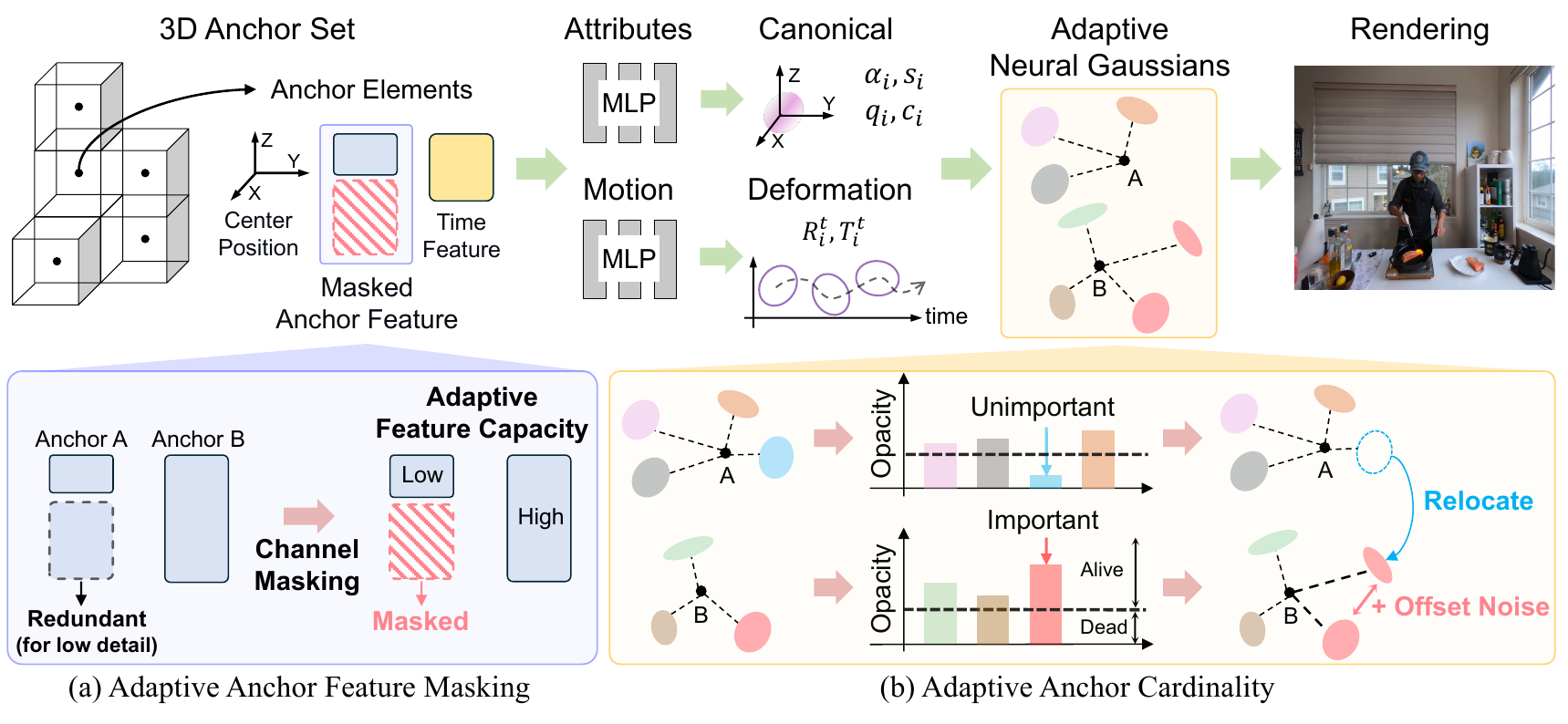}
    \vspace{-1em}
    \caption{Overall structure of our method. Given a set of 3D anchors, MLPs decode per-anchor and time-dependent features to generate Neural Gaussians for rendering. (a) Learnable masks eliminate representational redundancy for adaptive per-anchor capacity. (b) Low-importance Neural Gaussians are reallocated to high-importance regions based on accumulated importance, reducing redundant primitives while improving rendering quality.}
\label{fig:method}
\end{figure*}

\section{Related Work}
3D Gaussian Splatting (3DGS)~\cite{kerbl3Dgaussians} represents scenes with anisotropic 3D Gaussians and supports efficient rasterization and differentiable optimization, enabling 1080p rendering at over 30 FPS. While vanilla 3DGS relies on Structure-from-Motion (SfM)~\cite{sfm} initialization and adaptive densification, recent advancements have addressed its limitations in aliasing and geometric dependency. Notably, Mip-Splatting~\cite{Yu2023MipSplatting} introduces frequency-limited filters to mitigate sampling artifacts, while Scaffold-GS~\cite{scaffoldgs} adopts anchor-based hierarchical attributes to reduce redundancy and enforce structural consistency. Although these methods reduce redundancy and improve structural consistency, the explicit Gaussian attributes still incur substantial storage overhead, often exceeding gigabytes for complex scenes, thereby necessitating the development of effective compression techniques.

% Head 3
\subsection{Compression of 3DGS}

Early approaches reduce the primitive budget and attribute size through 
significance-based pruning, learnable masking, and attribute 
quantization~\cite{fan2023lightgaussian, lee2024c3dgs, girish2024eagles}. 
Other methods reorganize Gaussian attributes into compression friendly 
structures~\cite{morgenstern2024compact} or jointly optimize the 
representation under rate-distortion constraints~\cite{wang2024rdogaussian}. Recently, HAC~\cite{hac2024} and HAC++~\cite{hac++2025} demonstrated state-of-the-art bitrate efficiency by leveraging spatial consistency through hash-grid context models and anchor-based neural prediction, respectively. Additionally, OMG3D~\cite{lee2025omg} introduced Sub-Vector Quantization (SVQ) to balance codebook size and reconstruction fidelity, showing that exploiting local distinctiveness can drastically minimize the primitive budget while maintaining high rendering speeds. While these techniques have enabled 3DGS to represent static scenes efficiently, they are not directly applicable to dynamic scenes where scene content evolves over time, leading to new challenges in representation efficiency and scalability.

\subsection{3DGS for Dynamic Scenes}
Extending 3DGS to dynamic scenes typically follows two distinct paradigms: deformation-based and primitive-based approaches. Deformation-based methods~\cite{Wu_2024_CVPR,bae2024ed3dgs, yang2023deformable3dgs, Huang_2024_CVPR, li2025gifstream4dgaussianbasedimmersive, kratimenos2024dynmf} decouple the scene into a canonical 3D Gaussian field and a learnable deformation field parameterized by MLPs or grids. While storage-efficient, these methods often struggle to capture rapid motions or topological changes due to the limited capacity of the deformation network. Conversely, 4D primitive-based approaches~\cite{Li_STG_2024_CVPR, yang2023gs4d, yan20244d} explicitly model spatio-temporal Gaussians, enabling high-fidelity reconstruction of complex dynamic scenes. However, expanding primitives into the temporal dimension results in substantial memory consumption and lacks explicit temporal correspondence, hindering efficient compression. Recent hybrid approaches, such as Anchored 4DGS~\cite{10.1145/3757377.3763898} and 4D Scaffold~\cite{10.1609/aaai.v40i5.37332}, bridge this gap by utilizing sparse anchor points to generate local dynamic Gaussians, offering a balance between representational capacity and structural coherence suitable for compression. These advances in dynamic 3DGS representation naturally raise the question of how to effectively compress spatio-temporal Gaussian primitives while preserving visual fidelity and temporal coherence.

\subsection{Compression of 3DGS for Dynamic Scenes}
Compressing dynamic 3DGS presents unique challenges due to the compounding redundancy across both spatial and temporal dimensions. For primitive-based models, 4DGS-1K~\cite{yuan20251000fps4dgaussian} focused on lifespan-based pruning to eliminate transient primitives that contribute minimally to rendering. Also, CSTG~\cite{Lee_2024_C3DGS} used vector quantization and learnable masking methods to compress 4D Gaussian primitives in STG~\cite{Li_STG_2024_CVPR}. MEGA~\cite{zhang2024mega} further enhanced compactness by decoupling color into a DC-AC representation and employing entropy-constrained deformation to minimize the total Gaussian count. More recently, OMG4D~\cite{lee2025optimizedminimal4dgaussian} achieved extreme compaction by iteratively merging spatio-temporally adjacent Gaussians and applying 4D Sub-Vector Quantization (SVQ). For deformation-based methods, 4DGS-CC~\cite{chen20254dgscccontextualcodingframework} and Light4GS~\cite{liu2025light4gslightweightcompact4d} represented the deformation field using learnable context entropy coding. 

Recently, several works have compressed the anchor-based representation for dynamic scenes.
ADC-GS~\cite{ijcai2025p132} groups Gaussians into compact canonical anchors and proposes an anchor-level coarse-to-fine deformation with rate-distortion optimization to reduce deformation redundancy.
GIFStream~\cite{li2025gifstream4dgaussianbasedimmersive} extends the anchor representation with a feature stream for temporal prediction and employs an end-to-end compression pipeline with auto-regressive entropy coding for anchor attributes, while pruning time-dependent features for compact representation.
However, these anchor-based methods share an inherent limitation in structural flexibility, which can lead to suboptimal efficiency.
In this work, we propose an adaptive anchor representation that dynamically adjusts both the number of Neural Gaussians and the feature capacity per anchor, allocating higher representational capacity to fine-grained dynamic regions while reducing memory usage in less informative areas, thereby achieving higher visual fidelity with a lower overall memory footprint for dynamic scenes.

\section{Method}
In conventional anchor-based representations, high-complexity regions are handled by densifying anchors. However, a fixed number of Neural Gaussians and uniform feature capacity per anchor often lead to a substantially larger anchor set to achieve competitive quality.
We relax this constraint with two novel mechanisms that provide greater freedom in capacity allocation across both space and time while preserving compact, high-quality representations. 

Adaptive Anchor Cardinality (Section~\ref{sec:NGR}) enables anchors to generate a variable number of Neural Gaussians, concentrating primitives in regions with high-frequency geometric structures, while reducing redundancy in smooth regions.
Furthermore, this flexibility enables the use of fewer anchors in static regions by assigning more Neural Gaussians to each anchor, while focusing anchor density on dynamic areas that require many anchors for high-fidelity motion representation.

Adaptive Anchor Feature Masking (Section~\ref{sec:AFC}) further enhances representation flexibility by modulating feature dimensionality at the anchor level, assigning richer features to high-complexity regions and lightweight features to simpler ones.
In this approach, feature reduction on motion-critical anchors can introduce spatio-temporal interference, where capacity needed for temporal modeling is consumed by spatial representation. In Section~\ref{sec:opt}, we introduce a conditional regularization loss to control the reduction strength and preserve sufficient feature capacity in high-motion regions, and we present the complete training objective.

\noindent\textbf{Baseline Architecture.} We build on an anchor-based representation for dynamic scene modeling~\cite{scaffoldgs, li2025gifstream4dgaussianbasedimmersive}. Each voxelized anchor $i$, together with its associated offsets $\mathbf{o}_i\in\mathbb{R}^{K\times 3}$, generates $K$ local Neural Gaussians via anchor features $\mathbf{F}_i\in\mathbb{R}^{C_F}$ (Fig.~\ref{fig:method}), where $C_F$ denotes the feature channel dimension. To capture temporal variation, each anchor is further augmented with time-dependent features $\mathbf{T}_i\in\mathbb{R}^{T\times C_T}$, spanning $T$ video frames. 
Here, we learn to identify and mask static anchors that do not require motion representation using a masking parameter $d_i \in \mathbb{R}$, which is thresholded to either preserve or mask the time-dependent feature $\mathbf{T}_i$.
At each time step $t$, the network jointly consumes the static anchor feature $\mathbf{F}_i$ and the dynamic feature $\mathbf{T}_{i, t}\in\mathbb{R}^{C_T}$ to predict the spatial attributes of $K$ Neural Gaussians. These include color, rotation, scaling factor and opacity vectors, along with time-varying motion parameters that govern Gaussian deformation.

\subsection{Adaptive Anchor Cardinality}\label{sec:NGR}
As the first adaptive mechanism, we introduce \textit{Adaptive Anchor Cardinality} (AAC) to break the constraint of assigning a fixed number of Neural Gaussians per anchor. AAC enables heterogeneous 
per-anchor cardinality by using accumulated view-dependent opacity as a global importance score. Based on this score, we relocate less influential Gaussians from over-represented regions to anchors requiring higher primitive density. This redistribution enhances local structure without introducing additional anchor points, avoiding extra memory overhead (Fig.~\ref{fig:method}(b)).

\begin{figure}[t]
    \centering
    \includegraphics[width=0.85\linewidth]{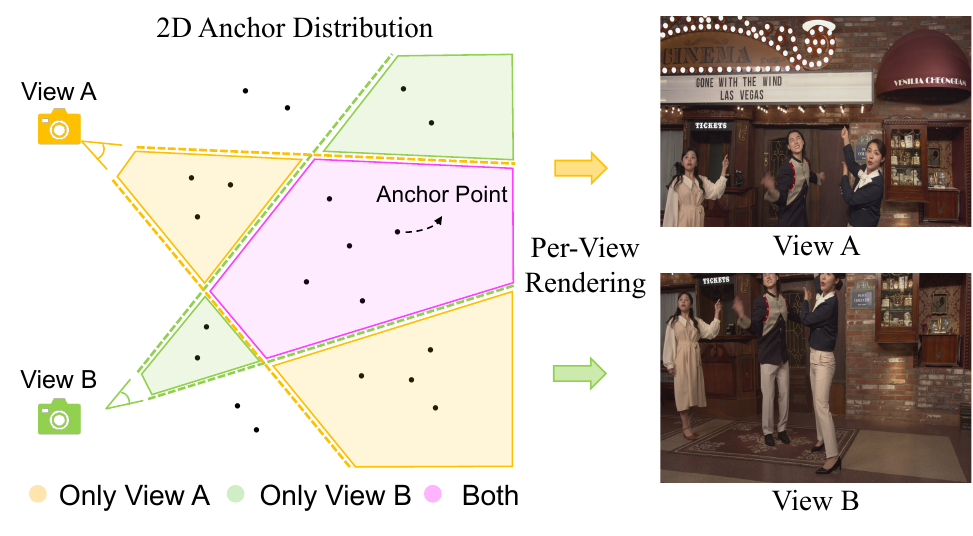}
    \vspace{-1em}
    \caption{Per-view visibility of anchors. Since only anchors within the view frustum are rendered, we accumulate opacity across views to estimate view-independent importance.}
\label{fig:per-view}
\vspace{-1em}
\end{figure}

\noindent \textbf{Importance of Neural Gaussians.} 
Relocation operates by identifying and moving less important Neural Gaussians, which requires a reliable measure of primitive importance. In 3DGS-based methods without anchor structure, opacity is commonly used as a proxy~\cite{kerbl3Dgaussians, kheradmand20243d}, where each Gaussian carries an explicit opacity attribute that can be evaluated independently of anchor visibility. In contrast, anchor-based Gaussians do not store per-primitive opacity directly: their view-dependent opacity is decoded on-the-fly by an MLP conditioned on the anchor features, and thus becomes available only when the parent anchor is spawned within the current camera’s view frustum (Fig.~\ref{fig:per-view}). Consequently, primitives that are important from other viewpoints can be temporarily unspawned and never decoded, receiving negligible importance when judged solely by instantaneous opacity.
To address this bias, we accumulate the opacity contributions of each Neural Gaussian over a relocation interval $T_R$ of iterations.
The resulting importance score $\mathcal{I}_{i,k}$ is defined with each training view $v_t$ at iteration $t$:
\begin{equation}
    \mathcal{I}_{i,k} = \sum_{t=1}^{T_R} \alpha_{i,k}^{(t)}(v_t),
\end{equation}
where $\alpha_{i,k}^{(t)}(\cdot)$ denotes the view-dependent opacity of each Neural Gaussian at iteration $t$.
By aggregating evidence across viewpoints and timestamps, this metric captures the global relevance of each dynamic primitive, independent of momentary view frustum and the corresponding temporal state. The score $\mathcal{I}_{i,k}$ is reset after each relocation step, ensuring that every relocation cycle is driven solely by newly observed contributions.

\noindent \textbf{Neural Gaussian Relocation.}
Guided by the accumulated importance scores $\mathcal{I}$, the relocation stage determines which Neural Gaussians to select and where to redistribute representational capacity. Rather than spawning additional anchors, it operates through a two-step procedure: importance-driven deactivation followed by sampling-based relocation.

In the first step, Neural Gaussians are ranked by their accumulated importance and split by a threshold $\tau$. Gaussians above $\tau$ are retained as active (alive), while those below $\tau$ are flagged as inactive (dead) and become candidates for relocation. To maintain training stability, we cap the number of selected inactive Gaussians to at most $\rho$\% of the total population per relocation step. Unless otherwise noted, we set $\tau = 0.005$ and $\rho = 20$ in our experiments.

In the second step, we relocate the selected inactive Neural Gaussians to anchors that require higher primitive density while preserving the total population. For each inactive Gaussian, we sample a parent Gaussian from the active set via importance-proportional sampling to determine its relocation target. In practice, since the sampling probability is governed by importance scores, active Gaussians in anchors covering fine-detail regions are significantly more likely to be selected as target anchors.
Through relocation, inactive Gaussians are reactivated by inheriting their parent’s attributes, including anchor membership. Their spatial positions are initialized by perturbing the parent offset $\mathbf{o}_{\text{parent}}$ with a perturbation $\boldsymbol{\epsilon}\in\mathbb{R}^3$ to avoid exact overlap while preserving local structure. Specifically, the offset of relocated Gaussian $\mathbf{o}_{\text{new}}$ is defined as follows,
\begin{equation}
    \mathbf{o}_{\text{new}} = \mathbf{o}_{\text{parent}} + |\mathbf{o}_{\text{parent}}| \odot \boldsymbol{\epsilon}, \,\, \boldsymbol{\epsilon} \sim \mathcal{N}(0, \delta^2 I),
\end{equation}
where $\delta$ controls the perturbation magnitude and is set to 0.05 by default, $|\cdot|$ denotes the element-wise absolute value, and $\odot$ denotes the Hadamard product.

\begin{table*}[t] 
\centering
\small
\begin{minipage}{0.53\textwidth}
    \centering
    \captionof{table}{Quantitative results on MPEG dataset. We evaluate our method on the Bartender and Cinema without image resolution downscaling. Storage size is measured in megabytes (MB). We highlight the \colorbox{red!20}{best} and \colorbox{orange!20}{second-best} performances.}
    \vspace{-0.7em}
    \label{tab:results_mpeg_split}
    \resizebox{0.9\linewidth}{!}{
    \begin{tabular}{lcccc}
    \toprule
    \multirow{2}{*}{Method} & \multicolumn{4}{c}{MPEG} \\ \cmidrule(lr){2-5}
     & PSNR↑ & SSIM↑ & LPIPS(VGG)↓ & Size↓ \\ \midrule
    4DGaussian~\cite{Wu_2024_CVPR} & 28.60 & 0.856 & 0.260 & 110  \\
    4DGS~\cite{yang2023gs4d} & 29.03 & 0.879 & \cellcolor{red!20}0.195 & 366.6 \\
    E-D3DGS~\cite{bae2024ed3dgs} & 28.01 & 0.861 & 0.247 & 35.6  \\
    STG~\cite{Li_STG_2024_CVPR} & 28.41 & 0.878 & \cellcolor{orange!20}0.217 & 60.8\\
    CSTG+PP~\cite{Lee_2024_C3DGS} & 27.50 & 0.867 & 0.233 & 9.7 \\
    GIFStream~\cite{li2025gifstream4dgaussianbasedimmersive} & \cellcolor{orange!20}29.89 & \cellcolor{red!20}0.900 & 0.223 & \cellcolor{orange!20}5.4 \\ \midrule
    \textbf{Ours} & \cellcolor{red!20}29.92 & \cellcolor{orange!20}0.898 & 0.228 & \cellcolor{red!20}3.5 \\ \bottomrule
    \end{tabular}
    }
\end{minipage}
\hfill 
\begin{minipage}{0.44\textwidth}
    \centering
    \includegraphics[width=0.86\linewidth]{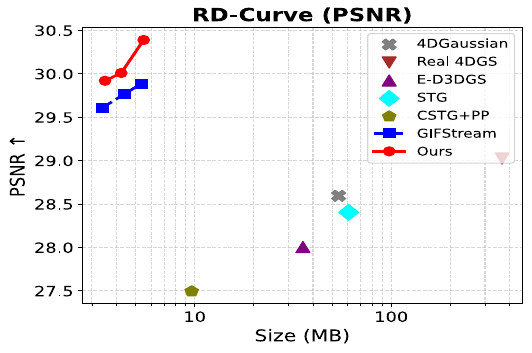}
    \vspace{-1em}
    \captionof{figure}{Rate-distortion (PSNR) curve on MPEG datasets. We control the model size exclusively via anchor densification for both GIFStream~\cite{li2025gifstream4dgaussianbasedimmersive} and Ours.}
    \label{fig:rd-curve}
\end{minipage}

\end{table*}

\noindent \textbf{Cardinality Gating Mask.} 
Allowing each anchor to host a varying number of Neural Gaussians poses a key architectural challenge. A naive design would require cardinality-specific decoding, for example, separate MLP decoders for different per-anchor Gaussian counts, which introduces heterogeneous tensor shapes across anchors. This could lead to excessive parameter growth and prevent efficient batch-parallel execution. We address this challenge with a per-Gaussian gating strategy, which allocates a fixed upper bound on the number of Gaussians per anchor and activates only the required subset, thereby decoupling the attribute prediction process from the actual Gaussian activation. In this design, the shared MLPs consistently predict attributes for a fixed maximum number of Neural Gaussians per anchor, regardless of how many are actually needed. And then, the decision of which Gaussians to activate is handled separately by the gating mask, allowing the network architecture to remain fixed and batch-friendly while the number of rendered primitives varies per anchor. 

Specifically, each gating value $g_{i,k} \in \{0, 1\}$ controls whether the $k$-th Neural Gaussian associated with anchor $i$ is active. Consistent with prior anchor-based designs, all Gaussian attributes are densely predicted by shared MLPs, independent of activation, rather than being conditionally decoded. Activation is instead handled at the rendering stage, with the gating mask applied at the opacity level, yielding the gated opacity $g_{i,k}\alpha_{i,k}$.
This leads to Neural Gaussians with $g_{i,k}=0$ being excluded from rendering, ignored prior to rasterization. This strategy enables anchors to flexibly adjust their active primitive counts while preserving a fixed-size, batch-parallel processing pipeline.

\begin{table*}[t]
\centering
\caption{Quantitative results on Panoptic Sports~\cite{joo2016panopticstudiomassivelymultiview} and N3DV~\cite{li2022neural3dvideosynthesis} datasets. Since we could not find configuration about Panoptic Sports (basketball, boxes), we reported the paper results of GIFStream~\cite{li2025gifstream4dgaussianbasedimmersive}.  Consistent with prior studies, images from N3DV are downscaled by a factor of 0.5. Storage size is measured in megabytes (MB).}
\label{tab:results_panoptic_n3dv}
\vspace{-1em}
% \small
% \setlength{\tabcolsep}{4.5pt} % 열이 많으므로 간격을 조금 더 좁게 설정
\resizebox{0.8\linewidth}{!}{\begin{tabular}{lcccccccc}
\toprule
\multirow{2}{*}{Method} & \multicolumn{4}{c}{Panoptic Sports} & \multicolumn{4}{c}{N3DV} \\ \cmidrule(lr){2-5} \cmidrule(lr){6-9}
 & PSNR↑ & SSIM↑ & LPIPS(VGG)↓ & Size↓ & PSNR↑ & SSIM↑ & LPIPS(ALEX)↓ & Size↓ \\ \midrule
Dynamic 3DGS~\cite{luiten2023dynamic} & 28.84 & 0.910 & 0.175 & 2083.8 & - & - & - & - \\
4DGaussian~\cite{Wu_2024_CVPR} & 27.71 & 0.914 & 0.175 & 137.7 & 31.15 & - & 0.049 & 90 \\
4DGS~\cite{yang2023gs4d} & 28.68 & 0.911 & 0.157 & 973.8 & \cellcolor{orange!20}32.01 & - & - & 202 \\
E-D3DGS~\cite{bae2024ed3dgs} & 25.61 & 0.896 & 0.172 & 297.9 & 31.42 & \cellcolor{orange!20}0.945 & \cellcolor{red!20}0.037 & 137 \\
STG~\cite{Li_STG_2024_CVPR} & 25.09 & 0.900 & 0.181 & 180.9 & \cellcolor{red!20}32.05 & \cellcolor{red!20}0.946 & \cellcolor{orange!20}0.044 & 200 \\
CSTG+PP~\cite{Lee_2024_C3DGS} & 26.13 & 0.902 & 0.192 & 23.4 & 31.69 & \cellcolor{orange!20}0.945 & 0.054 & 15 \\
GIFStream~\cite{li2025gifstream4dgaussianbasedimmersive} & \cellcolor{red!20}29.50 & \cellcolor{red!20}0.931 & \cellcolor{red!20}0.114 & \cellcolor{orange!20}12.6 & 31.75 & 0.938 & 0.051 & \cellcolor{orange!20}10 \\ \midrule
\textbf{Ours} & \cellcolor{orange!20}29.30 & \cellcolor{orange!20}0.922 & \cellcolor{orange!20}0.128 & \cellcolor{red!20}7.3 & 31.73 & 0.940 & 0.058 & \cellcolor{red!20}6.3 \\ \bottomrule
\end{tabular}}
\end{table*}

\subsection{Adaptive Anchor Feature Masking}
\label{sec:AFC}
AAC improves memory efficiency by refining the spatial allocation of Neural Gaussians using a compact set of primitives. However, redundancy may still persist in the feature channels assigned to each anchor, as a fixed feature dimensionality is imposed despite the diverse representational demands across different regions. To address this inefficiency, \textit{Adaptive Anchor Feature Masking} (AAFM) dynamically modulates the number of active feature channels per anchor using learnable masks (Fig.~\ref{fig:method}(a)).

While aggressive sparsification is desirable for compactness, empirical evidence indicates that treating all channels as elimination candidates can destabilize optimization. To address this, each anchor feature $\mathbf{F}_i$ is factorized into a base segment $\mathbf{F}_i^{\text{base}}$ and an adaptive segment $\mathbf{F}_i^{\text{adapt}}$. AAFM restricts masking exclusively to $\mathbf{F}^{\text{adapt}}$, while keeping the base segment $\mathbf{F}^{\text{base}}$ fully active throughout training. The base component thus functions as a stable representational backbone, preserving essential capacity during learning.
Accordingly, we adopt an additional mask parameter $m_i\in\mathbb{R}$ for each anchor $i$ to learn a binary mask $M_i\in\{0,1\}$ via the 
straight-through estimator (STE)~\cite{bengio2013estimatingpropagatinggradientsstochastic}, and reparameterize the anchor feature as follows:
\begin{equation}
    \begin{split}
    \mathbf{F}_i &= \text{concat}(\mathbf{F}_i^{\text{base}}, M_i\mathbf{F}_i^{\text{adapt}}), \\
    M_i &= \operatorname*{sg}(\mathbf{1}[\sigma(m_i) > \gamma] - \sigma(m_i)) + \sigma(m_i),
    \end{split}
\end{equation}
where $\text{concat}(\cdot,\cdot)$ denotes concatenation along the feature channel dimension, $\gamma$ is the masking threshold, and $\text{sg}(\cdot)$, $\mathbf{1}[\cdot]$, and $ \sigma(\cdot)$ are the stop-gradient, indicator, and sigmoid functions, respectively.

Under this design, feature sparsity is encouraged via a mask regularization term, $\mathcal{L}_{\text{M}}$, which penalizes the activation of adaptive feature channels and is formulated as follows:
\begin{equation}
    \mathcal{L}_{\text{M}} =\frac{1}{N} \sum_{i=1}^{N} \sigma(m_i),
\end{equation}
where $\sigma(\cdot)$ is the sigmoid function, and $N$ is the total number of anchors.
When the reconstruction loss indicates that $\mathbf{F}^{\text{base}}$ alone is sufficient, the corresponding masks of $\mathbf{F}^{\text{adapt}}$ naturally converge toward deactivation, suppressing redundant feature channels. Rather than explicitly pruning these channels, which would induce varying input dimensionalities for subsequent MLPs, masking is applied via element-wise modulation after training to preserve a fixed input shape. The actual removal of deactivated channels is reserved for the post-training stage, with the balance between reconstruction fidelity and feature sparsity.

\subsection{Optimization}\label{sec:opt}

\textbf{Conditional Regularization.} A naive integration of AAFM often leads to functional overlap between spatial and temporal features. To maintain spatial quality, time-dependent features tend to be used to compensate for the reduced expressivity caused by anchor feature masking. This capacity leakage distorts the intended role of temporal features and degrades compression efficiency by inhibiting effective sparsification. 

To mitigate such behavior, we introduce a \textit{Conditional Regularization Loss} ($\mathcal{L}_{\text{CR}}$) that regulates anchor feature sparsification based on local temporal activity. This loss drives feature masking in regions where temporal features are weak or inactive, typically corresponding to static or simple areas, while suppressing sparsification in temporally dynamic regions. We can formulate the conditional regularization loss as follows: 
\begin{equation}
    \mathcal{L}_{\text{CR}} = \frac{1}{N} \sum_{i=1}^{N} \sigma(d_i) (1 - \sigma(m_i)),
\end{equation}
where $\sigma(\cdot)$ is the sigmoid function, and $m_i, d_i$ denote the anchor and time-dependent feature masks.
By constraining where anchor feature capacity can be reduced, this design prevents time-dependent features from absorbing loss of spatial details, thereby preserving their dedicated role in temporal modeling.

\noindent\textbf{Training Objective.} We optimize the entire model in an end-to-end manner, and the final objective $\mathcal{L}$ is defined as follows:
\begin{equation}
\mathcal{L} = \mathcal{L}_{\text{base}} + \lambda_{\text{M}}\mathcal{L}_{\text{M}} + \lambda_{\text{CR}}\mathcal{L}_{\text{CR}}, %+ \lambda_{\text{NC}}\mathcal{L}_{\text{NC}},
\end{equation}
where $\mathcal{L}_{\text{base}}$ includes the rendering loss, entropy loss for compression, and regularization terms used in GIFStream~\cite{li2025gifstream4dgaussianbasedimmersive}.

\section{Experiments}
We evaluate our method on three real-world multi-view video datasets: N3DV~\cite{li2022neural3dvideosynthesis}, MPEG, and Panoptic Sports~\cite{joo2016panopticstudiomassivelymultiview}. For quantitative evaluation, we report PSNR, SSIM~\cite{wang2004ssim}, 
and LPIPS~\cite{zhang2018perceptual}, along with storage size in 
megabytes (MB). Following prior work, LPIPS is computed with AlexNet 
on N3DV and with VGG on the other datasets.
We implement our model on top of the GIFStream framework~\cite{li2025gifstream4dgaussianbasedimmersive}, maintaining the original voxelized anchor initialization and densification strategy. For stable training, AAC is activated every 500 iterations starting from 3,000 iteration, once anchor densification has stabilized. In contrast, AAFM is active from the start, with the regularization weight $\lambda_{\text{M}}$ tuned to reach a final masking ratio of about 60\%. After training, as a post-hoc step, we prune deactivated Neural Gaussians and masked anchor features.

\subsection{Results}
\begin{figure}[t]
    \centering
    \includegraphics[width=0.9\linewidth]{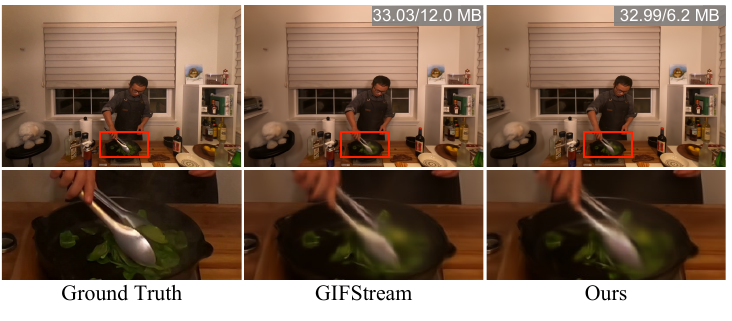}
    \vspace{-1em}
    \caption{Qualitative comparison on N3DV Cook Spinach~\cite{li2022neural3dvideosynthesis} (300 frames, 1352 $\times$ 1014).}
\label{fig:cook_ql}
\vspace{-1em}
\end{figure}

As shown in Tables~\ref{tab:results_mpeg_split} and \ref{tab:results_panoptic_n3dv}, our method attains the most compact representation across all evaluated benchmarks. Compared to uncompressed baselines such as 4DGS and STG that incur substantial storage overhead, our approach achieves a significant reduction in memory footprint while providing superior or comparable reconstruction fidelity. Specifically, our model maintains an efficient footprint of roughly 3.5 to 7.3~MB across all scenes, consistently outperforming these uncompressed references in both efficiency and quality. Furthermore, our method remains superior to heavily compressed baselines like CSTG+PP across nearly all quality metrics. Notably, compared to the state-of-the-art anchor-based GIFStream, our framework achieves a further 35\% to 42\% reduction in storage while matching its visual performance (Fig.~\ref{fig:cook_ql}). The rate-distortion curves on the MPEG datasets in Fig.~\ref{fig:rd-curve} further highlight this advantage, as our method consistently attains superior quality metrics compared to GIFStream across the entire evaluated size range. These quantitative evaluations confirm that our adaptive-capacity design effectively removes spatiotemporal redundancy, ensuring high-fidelity rendering across diverse dynamic scenes without excessive memory overhead.

\begin{table}[t]

\centering

\caption{Ablation study of each module of the proposed method where AAC, AAFM, Comp. refer to Adaptive Anchor Cardinality, Adaptive Anchor Feature Masking and end-to-end compression strategy of GIFStream~\cite{li2025gifstream4dgaussianbasedimmersive}.} 

\label{tab:quantitative_ab}
\vspace{-1em}
\resizebox{0.95\linewidth}{!}{

\begin{tabular}{c|cccccccc}

\toprule

\multicolumn{3}{c}{Method}                      & \multicolumn{6}{c}{MPEG Bartender}        \\\cmidrule(lr){1-3}\cmidrule(lr){4-9} 

Comp.             & AAC                  & AAFM

& PSNR  & SSIM  & LPIPS & FPS & \# Anchors & Size(MB) \\ \midrule

\multirow{3}{*}{--} & -- & --  & 32.07 & 0.900 & 0.226 & 55 &111,759    & 248.3  \\

        &      \checkmark        &       --       & 32.30 & 0.901 & 0.225 & 67 & 59,336     & 132.2 \\

& \checkmark  & \multicolumn{1}{c}{\checkmark} & 32.13 & 0.898 & 0.234 & 71 & 54,828     &117.6\\ 

\cmidrule(lr){1-3}\cmidrule(lr){4-9}

\multirow{3}{*}{\checkmark} & \multicolumn{2}{c}{GIFStream}  & 31.94 & 0.898 & 0.227 & 77 &111,759    & 5.3  \\

        &      \checkmark        &       --       & 32.12 & 0.899 & 0.227 & 83 & 59,336     & 4.0 \\

& \checkmark  & \multicolumn{1}{c}{\checkmark} & 32.02 & 0.896 & 0.235 & 84 & 54,828     &3.5\\ 

\bottomrule
\end{tabular}}
\vspace{-0.5em}
\end{table}

\begin{table}[t]
\centering
\caption{Ablation study of Adaptive Anchor Cardinality (AAC). Under adaptivity, cardinality denotes the maximum Neural Gaussian count per anchor.} 
\vspace{-1em}
\label{tab:quantitative_ab_AAC}
\resizebox{1.0\linewidth}{!}{
\begin{tabular}{c|cccccccc}
\toprule

Adaptivity & \multicolumn{2}{c}{Cardinality}  & PSNR  & SSIM  & LPIPS & \# Gaussians & \# Anchors & Size(MB) \\ \cmidrule(lr){1-3} \cmidrule(lr){4-9}

-- & \multicolumn{2}{c}{5}  & 32.07 & 0.900 & 0.226 & 558,795 & 111,759    & 248  \\\midrule

\checkmark & \multicolumn{2}{c}{6}  & 31.88 & 0.895 & 0.237 & 196,501 & 58,899     & 130 \\

% \checkmark & \multicolumn{2}{c}{10 (Ours)}  & 32.13 & 0.898 & 0.234 & 250,792 & 54,828     &118 \\ 
\textbf{\checkmark} & \multicolumn{2}{c}{\textbf{10 (Ours)}} & \textbf{32.13} & \textbf{0.898} & \textbf{0.234} & \textbf{250,792} & \textbf{54,828} & \textbf{118} \\ 

\checkmark & \multicolumn{2}{c}{14}  & 32.02 & 0.896 & 0.233 & 267,225 & 52,905    & 122 \\

\bottomrule
\end{tabular}}
\vspace{-1em}
\end{table}

\begin{table}[t]
\centering
\caption{Ablation study of Adaptive Anchor Feature Masking (AAFM). 'Total CH' denotes the total dimension of an anchor feature, while $\mathbf{F}^{\text{base}}$ and $\mathbf{F}^{\text{adapt}}$ represent the number of fixed and maskable channels, respectively.} 
\label{tab:quantitative_ab_AAFM}
\vspace{-1em}
\resizebox{0.9\linewidth}{!}{
\begin{tabular}{c|cccccc}
\toprule
\multicolumn{7}{c}{N3DV Flame Steak}        \\\midrule 
Total CH & \multicolumn{2}{c}{$\mathbf{F}^{\text{base}}$ | $\mathbf{F}^{\text{adapt}}$}  & PSNR  & SSIM  & LPIPS(ALEX) & Size(MB) \\ \cmidrule(lr){1-3} \cmidrule(lr){4-7}

24 & \multicolumn{2}{c}{12\hspace{5pt}|\hspace{5pt}12}  & 32.96 & 0.953 & 0.044 & 5.34 \\

\textbf{48 (Ours)} & \multicolumn{2}{c}{\textbf{24\hspace{5pt}|\hspace{5pt}24}} & \textbf{33.64} & \textbf{0.956} & \textbf{0.043} & \textbf{6.15} \\ 

64 & \multicolumn{2}{c}{32\hspace{5pt}|\hspace{5pt}32}  & 33.57 & 0.957 & 0.041 & 6.77 \\
\bottomrule
\end{tabular}}
\vspace{-0.5em}
\end{table}

\begin{table}[t]
\caption{Ablation study on the anchor feature masking regularization weight $\lambda_{\text{M}}$.}
\label{tab:aafm_weight}\vspace{-1em}
\resizebox{0.8\linewidth}{!}{
\begin{tabular}{lccccc}
\toprule
$\lambda_{\text{M}}$    & Masking Ratio & PSNR & SSIM &LPIPS &Size (MB) \\
\midrule
3e-3      &  32\%   & 32.07 & 0.897 & 0.232 & 3.8     \\
\textbf{8e-3}      &  61\%   & 32.02 & 0.896 & 0.235 & 3.5       \\
3e-2       &  89\%   & 31.50 & 0.890 & 0.241 & 3.0       
\\\bottomrule
\end{tabular}}
\vspace{-0.5em}
\end{table}

\begin{table}[t]
\centering
\caption{Ablation studies on $\lambda_{\text{CR}}$, $\gamma$, $\tau$ and $\rho$.}
\label{tab:hyperparam_ablation}\vspace{-1em}
\resizebox{0.8\linewidth}{!}{
\begin{tabular}{c|c|cccc}
\toprule
Param. & Value & PSNR & SSIM & LPIPS & Size (MB) \\
\midrule
\multirow{3}{*}{$\lambda_{\text{CR}}$}
       & 0.001  & 31.80 & 0.894 & 0.237 & 3.6 \\
       & \textbf{0.1 (Ours)}    & 32.02 & 0.896 & 0.235 & 3.5 \\
       & 0.2    & 32.01 & 0.897 & 0.234 & 3.5 \\
\midrule
\multirow{3}{*}{$\gamma$}
       & 0.001          & 31.84 & 0.894 & 0.236 & 3.5   \\
       & \textbf{0.01 (Ours)}   & 32.02 & 0.896 & 0.235 & 3.5   \\
       & 0.1            & 31.98 & 0.897 & 0.233 & 3.4   \\
\midrule
\multirow{3}{*}{$\tau$}
       & 0.0001 & 32.01     & 0.897     & 0.236     & 3.5   \\
       & \textbf{0.005 (Ours)}  & 32.02 & 0.896 & 0.235 & 3.5 \\
       & 0.01           & 32.03 & 0.897 & 0.236 & 3.6   \\
\midrule
\multirow{3}{*}{$\rho$}
       & 10\%            & 31.96     &  0.895    & 0.239     & 3.6   \\
       & \textbf{20\% (Ours)}  & 32.02 & 0.896 & 0.235 & 3.5 \\
       & 30\%           & 32.01 & 0.897 & 0.234 & 3.4  \\
\bottomrule
\end{tabular}}
\end{table}

\begin{figure*}[t]
    \centering
    \includegraphics[width=0.9\textwidth]{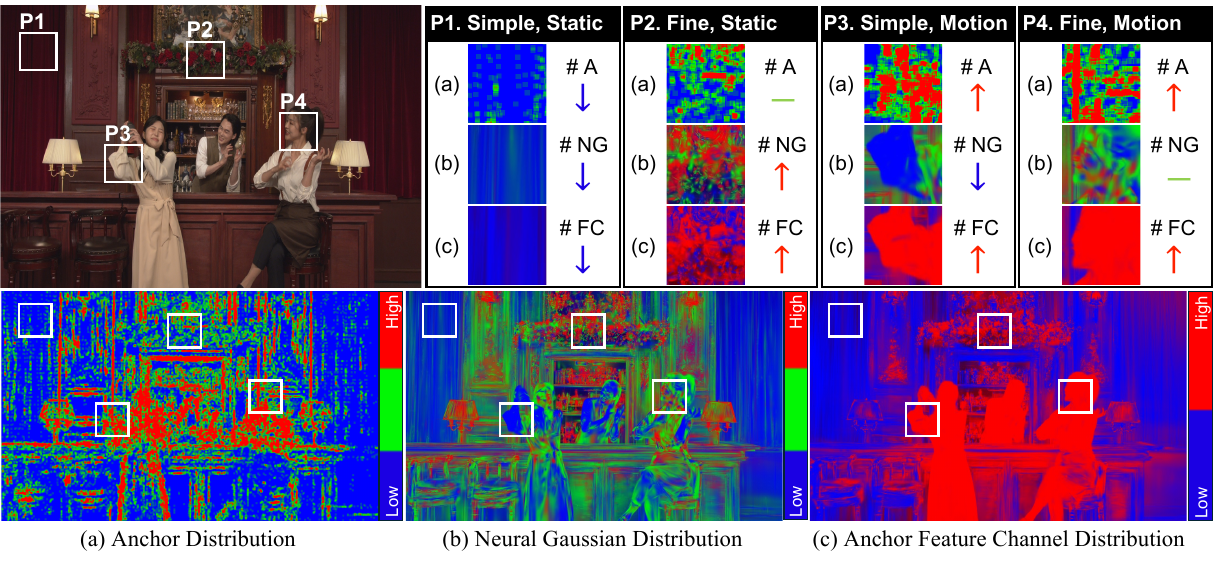}
    \vspace{-1em}
    \caption{Visualization. In this figure, we visualize the pixel-wise density of anchor components using a heatmap spectrum from Low (Blue) to High (Red). (a) Anchor Distribution: Blue, Green, and Red correspond to anchor counts of 0--1, 2--4, and $\ge$5 per pixel, respectively. (b) Neural Gaussian Distribution: The colors represent the cardinality of Neural Gaussians: 1--3 (Blue), 4--6 (Green), and 7--10 (Red). (c) Anchor Feature Channel Distribution: Blue denotes 50\% active channels, while Red denotes full channel usage. We provide a detailed analysis of each region in Section~\ref{sec:ablation}.}
\label{fig:visualization}
\end{figure*}

\subsection{Ablation Study}
\label{sec:ablation}

\noindent \textbf{Overall Architecture.} Table~\ref{tab:quantitative_ab} evaluates each module with and without the learned compression pipeline~\cite{li2025gifstream4dgaussianbasedimmersive}. Without compression, AAC alone reduces the anchor count from 111,759 to 59,336, while improving PSNR (32.07 to 32.30) and speed (55 to 67~FPS). Adding AAFM—trained with conditional regularization loss—further reduces the size to 117.6~MB and boosts speed to 71~FPS, demonstrating that primitive relocation and feature pruning synergistically optimize capacity. When integrated with the compression pipeline, our models further reduce storage by 33.9\% while maintaining quality. These results confirm that adaptive resource allocation effectively complements existing attribute compression methods.

Fig.~\ref{fig:visualization} illustrates the adaptive distribution of anchors, Neural Gaussians, and feature channels across four regions (P1--P4) with distinct spatiotemporal characteristics. In simple static areas (P1), the model prioritizes efficiency through minimal resource allocation, whereas detailed static regions (P2) leverage increased Gaussian counts and feature dimensions to enhance representational power. For dynamic regions, since motion representation relies on rigid transformations via anchor-level deformation, accurately capturing fast and large-scale motion requires increasing the number of anchors in both P3 and P4. Also, within these regions, P3 utilizes fewer Neural Gaussians given its simple spatial structure, whereas P4 allocates a sufficient number of primitives to capture the required fine-grained detail. Notably, Conditional Regularization maintains high feature capacity in P3 and P4, mitigating training difficulty in dynamic areas. These results demonstrate that our method adaptively scales resources based on local complexity and motion, achieving high quality while maintaining model compactness.

\noindent \textbf{Adaptive Anchor Cardinality.} Table~\ref{tab:quantitative_ab_AAC} evaluates the impact of the maximum cardinality setting within the AAC strategy. Our results indicate that limited cardinality constrains the expressivity of individual anchors, requiring more anchors to capture geometric details. Conversely, excessively high cardinality induces Gaussian over-concentration, which leads to floater artifacts and compromises rendering quality. Consequently, we adopt a maximum cardinality of 10 as the optimal trade-off between reconstruction fidelity and computational efficiency, ensuring robust generalization across various scene scales.

\noindent \textbf{Adaptive Anchor Feature Masking.} Table~\ref{tab:quantitative_ab_AAFM} evaluates anchor feature capacity on N3DV Flame Steak by partitioning it into fixed ($\mathbf{F}^{\text{base}}$) and maskable ($\mathbf{F}^{\text{adapt}}$) components. We observe a trade-off where lower channel capacity degrades reconstruction quality, while excessive dimensions offer diminishing PSNR gains. Our 48-channel configuration attains an optimal balance, delivering high fidelity within a compact footprint. AAFM thus effectively suppresses redundancy, enabling complex spatiotemporal modeling without the overhead of larger feature budgets. Table~\ref{tab:aafm_weight} further analyzes the regularization weight $\lambda_{\text{M}}$ on MPEG Bartender. An overly large value induces excessive masking and degrades quality. We therefore set $\lambda_{\text{M}}$ to reach a masking ratio of about 60\%, which consistently preserves quality across scenes.

\noindent \textbf{Hyperparameter Sensitivity.} Table~\ref{tab:hyperparam_ablation} reports sensitivity analyses for the main hyperparameters of AAC and AAFM. Performance remains stable across a reasonable range of values for all parameters, indicating that our method is robust to hyperparameter choices. We use the default settings across all experiments.

\section{Conclusions}
In this paper, we proposed an adaptive-capacity anchor-based framework for dynamic scenes, designed to overcome the structural rigidity inherent in conventional anchor-based methods. Our approach dynamically allocates anchor representational capacity based on spatiotemporal demands. To achieve this, Adaptive Anchor Cardinality relocates unnecessary Neural Gaussians to regions requiring higher geometric detail without introducing additional anchors. Furthermore, Adaptive Anchor Feature Masking reduces redundant anchor features via learnable channel masking with regularization for stable spatiotemporal modeling. Experimental results on MPEG, Panoptic Sports, and N3DV show that our method improves storage efficiency without sacrificing visual quality or real-time rendering, achieving notably higher compression on fast motion scenes. As a next step, we aim to scale our framework to long video sequences~\cite{xu2024longvolcap} by enhancing inter-frame consistency. Also, integrating a hierarchical anchor representation~\cite{ren2024octree} will further optimize compression by modulating complexity across spatial scales.

% \clearpage

\begin{acks}
This work was supported in part by the Institute of Information and Communications Technology Planning and Evaluation (IITP) grants (2018-0-00207, RS-2019-II190421, RS-2020-II201821, RS-2021-II212052, RS-2025-02217613, RS-2025-25442569) funded by the Korean government (MSIT); the National Research Foundation of Korea (NRF) grants (RS-2024-00345732, RS-2025-02216217) funded by the Korean government (MSIT); and the Technology Innovation Program (RS-2023-00235718, 23040-15FC, 1415187505) and the K-CHIPS (Korea Collaborative \& High-tech Initiative for Prospective Semiconductor Research) program (RS-2026-25521189, Development of Application Processor Design Technologies for Multimodal Sensor Signal Processing) funded by the Ministry of Trade, Industry and Energy (MOTIE, Korea).
\end{acks}

% Bibliography
\bibliographystyle{ACM-Reference-Format}
\bibliography{MM_bibliography}

\end{document}